\documentclass{article}

\PassOptionsToPackage{numbers, compress}{natbib}

\usepackage{enumitem}
\usepackage[main, final]{neurips_2026}

\usepackage[utf8]{inputenc}
\usepackage[T1]{fontenc}
\usepackage{hyperref}
\usepackage{url}
\usepackage{booktabs}
\usepackage{graphicx}
\usepackage{subcaption}
\usepackage{algorithm}
\usepackage{algpseudocode}
\usepackage{amsmath}
\usepackage{amsfonts}
\usepackage{placeins}
\usepackage{nicefrac}
\usepackage{microtype}
\usepackage{xcolor}
\usepackage{tikz}
\usepackage[capitalize,noabbrev]{cleveref}

\definecolor{cornflowerblue}{rgb}{0.39, 0.58, 0.93}
\hypersetup{
    colorlinks=true,
    linkcolor=cornflowerblue,
    filecolor=magenta,      
    urlcolor=teal,
    citecolor=cornflowerblue,
    pdftitle={sleeping},
    pdfpagemode=FullScreen,
}

\usetikzlibrary{arrows.meta}

\newcommand{\sleepPasses}{N}
\newcommand{\windowSize}{L}
\newcommand{\sequenceLen}{T}
\newcommand{\rolloutStep}{t}
\newcommand{\hopCount}{k}
\newcommand{\modelDepth}{D}

\newcommand{\vect}[1]{\boldsymbol{#1}}
\newcommand{\mat}[1]{\mathbf{#1}}
\newcommand{\R}{\mathbb{R}}

\newcommand{\q}{\vect{q}}
\newcommand{\kvec}{\vect{k}}
\renewcommand{\v}{\vect{v}}
\newcommand{\x}{\vect{x}}

\newcommand{\ovec}{\vect{o}}

\newcommand{\K}{\mat{K}}
\newcommand{\V}{\mat{V}}

\newcommand{\Sb}{\mat{S}}

\newcommand{\WQ}{\mat{W}_Q}
\newcommand{\WK}{\mat{W}_K}
\newcommand{\WV}{\mat{W}_V}

\DeclareMathOperator{\softmax}{softmax}

\title{Do Language Models Need Sleep?  \\
 Offline Recurrence for Improved Online Inference}

\author{%
  Sangyun Lee\thanks{Correspondence to: \texttt{sangyunl@andrew.cmu.edu}} \\
  Carnegie Mellon University \\
  \And
  Sean McLeish \\
  University of Maryland \\
  \And
  Tom Goldstein \\
  University of Maryland
  \And
  Giulia Fanti \\
  Carnegie Mellon University
}

\begin{document}

\maketitle

\begin{abstract}
  Transformer-based large language models are increasingly used for long-horizon tasks; however, their attention mechanism scales poorly with context length.
  To handle this, we study a sleep-like consolidation mechanism in which a model periodically converts recent context into persistent fast weights before clearing its key-value cache.
  During the sleep, the model performs $\sleepPasses$ offline recurrent passes over the accumulated context and updates the fast weights in its state-space model (SSM) blocks through a learned local rule.
  During inference, this shifts extra computation to the sleep while preserving the latency of wake-time prediction.
  We test our method on controlled synthetic tasks, including cellular automata and multi-hop graph retrieval, as well as a realistic math reasoning task, on which SSM-attention hybrid models fail.
  We then show that increasing sleep duration $\sleepPasses$ for our models improves performance, with the largest gains on examples that require deeper reasoning. Code available at https://github.com/sangyun884/llm-sleep
\end{abstract}

\section{Introduction}
Large Language Models (LLMs) are commonly based on the transformer architecture
\citep{vaswani2017attention}, which stores context in an attention cache and
retrieves past tokens as needed. This memory mechanism is central to their
performance, but it scales poorly: total attention compute grows quadratically
with context length, while cache memory grows linearly. 

Recent efficient sequence models~\cite{ren2024samba,dong2024hymba,de2024griffin,arora2024simple} mitigate this cost by introducing fixed-size fast weight memories~\cite{yang2024gated,dao2024transformers,schlag2021linear} interleaved with full self-attention. 
This hybrid design brings together two complementary forms of memory: attention for high-fidelity access to recent tokens, and weight-based memory for compressed information beyond the active context window.
Hybrid models are now common among large scale frontier models \citep{qwen35blog}. 

However, scalable memory is not the same as scalable reasoning.
A fast weight memory may support long-range recall~\cite{ren2024samba}, but it is unclear whether it can support deep computation over tokens that are no longer present in the KV cache.
We find that the performance of vanilla SSM-attention hybrid models degrades (under the same token budget) as the required reasoning depth increases \textbf{even when the amount of information to store is held fixed. }
This suggests that the bottleneck is not merely memory capacity as suggested by prior work~\cite{jelassi2024repeat,arora2024simple}, but the amount of computation available for transforming evicted context into a useful internal state.

\textbf{Sleep.}
In animals, the transfer from short-term memory to long-term memory is thought to be supported by hippocampal replay \cite{mcclelland1995there}, especially during sleep \cite{rasch2013sleep}; in this  phase, short-term hippocampal memories are reactivated and consolidated into cortical synaptic weights.
Sleep makes animals unable to respond to external stimuli, suggesting that it must provide enough cognitive benefit to justify this cost~\cite{rasch2013sleep}.
Inspired by these biological processes, we propose a method for transferring context-window memory into persistent weights.
When the model's context window becomes full during inference, the model enters a ``sleep'' in which it performs multiple forward passes over the accumulated context and recursively updates its fast weights via a learned local rule.
As in animal sleep, the model receives no external input tokens during this phase.
After consolidation, the context window is cleared, and the model resumes operation with updated fast weights.
During training, the model is optimized end-to-end by backpropagating through the entire process to maximize task performance after sleep.

Our architecture is also motivated by results on depth-recurrent or looped neural networks \cite{graves2016adaptive,dehghani2018universal,bai2019deep}.
Prior work shows that dynamic-depth models can outperform fixed-depth counterparts on sequential reasoning tasks and solve hard problem instances that fixed-depth models cannot
by scaling amount of compute spent at prediction.
\textbf{Our key insight is that recurrence can be used not only for prediction but also for memory consolidation.}
Converting observed tokens into useful weight memory is itself a nontrivial computation, and need not be achievable in a single pass.
Indeed, many learning algorithms, such as gradient descent, improve through iterative weight updates.
Thus, allocating more recurrent computation during fast weight formation gives the model more steps to transform context into representations that support later prediction.
We find that increasing the depth of recurrence, or \textit{sleep duration}, improves reasoning after sleep.
Unlike previous looped models, our model does not need to loop at prediction time: the additional computation has already been spent on forming fast weights that support later single-pass prediction.

We introduce and evaluate LLM sleep on carefully designed \textit{synthetic tasks} where a model must answer questions about context that has already been evicted, using only a single forward pass.
These synthetic tasks allow us to vary reasoning depth while holding memory load fixed,
providing a clean stress test of whether sleep-time computation can convert
transient context into fast weights that support later inference.
We summarize our contributions as follows:
\begin{itemize}[leftmargin=*]
    \item In a controlled setting, we show that as the reasoning depth of a problem increases, vanilla State-Space Models (SSMs) such as Gated Delta Nets (GDNs) fail despite having enough fast weight capacity.
    \item We propose an architecture that combines recurrent computation with fast weight memory blocks, and show that increasing the number of recursions for our architecture improves performance over GDNs. We observe the largest gains on problem instances that require the deepest reasoning.
    \item We further validate the efficacy of our architecture on GSM-Infinite, a natural language math-reasoning dataset, using pre-trained LLM initializations.
\end{itemize}
Overall, these results support the central claim that a sleep-like offline recurrence can organize evicted context into weights to support later reasoning.

\section{Related Work}
\label{sec:related-work}

\textbf{Fast weights and linear recurrent neural networks.}
Linear recurrent neural networks or SSMs can be viewed as maintaining an online fast weight memory rather than a KV cache which grows linearly with sequence length. 
In this view, linear attention corresponds to a recurrent update over a fixed-size, matrix-valued, state, where key-value mappings are written and queried \citep{katharopoulos2020transformers,schlag2021linear}. 
Recent variants improve this memory with delta-rule updates and gates, enabling more selective writing, overwriting, and forgetting \citep{yang2023gatedlinear,yang2024gated,yang2024parallelizing,dao2024transformers}.
These mechanisms underlie recent efficient hybrid language models \citep{gu2025jet,nvidia2025nemotronnano2} and help explain why linear networks can offer a favorable recall, throughput, and memory tradeoffs.
They still struggle with exact copying and retrieval relative to full attention in some cases due to a fixed memory size, as pointed out by prior work \citep{arora2024simple,jelassi2024repeat}.
Contrary to these works, we show that such models can fail as the required reasoning depth to solve a task increases, \emph{even when the amount of information to store is held fixed}.

\textbf{Context compression.}
There are several methods for processing long contexts at test time by condensing contextual information.
\citet{ge2023context} propose using a language model to compress long contexts into a shorter sequence of hidden states, which are then passed to the language model in place of the original long context.
\citet{eyuboglu2025cartridges} use offline self-study to learn a small KV cache that can substitute for the full-context cache.
This line of work shares our goal of spending offline computation once to turn a long context into a compact state that can be reused later.
These methods shorten what remains in the attention context, whereas our method transfers evicted context into weight-based memory.

\textbf{Context distillation.}
Context distillation~\citep{snell2022learningdistillingcontext,askell2021generallanguageassistantlaboratory} aims to distill active context into model weights by training a model without it to imitate a contextful
teacher~\citep{snell2022learningdistillingcontext,askell2021generallanguageassistantlaboratory,caccia2025trainingplugnplayknowledgemodules},
reconstruct it~\citep{chen2024generativeadaptercontextualizinglanguage},
predict its continuation~\citep{caccia2025trainingplugnplayknowledgemodules,chen2024generativeadaptercontextualizinglanguage},
or answer questions about it~\citep{tack2024onlineadaptationlanguagemodels,cao2025infiniteiclbreakinglimitcontext,caccia2025trainingplugnplayknowledgemodules}.
Instead of doing gradient descent on predefined losses, our method uses a learned recurrent forward pass to transfer context to weights.

\textbf{Test-time training.}
Test-time training methods such as \citet{zhang2026test,tandon2025end} replace full attention with sliding-window attention and perform test-time gradient updates on a subset of MLP layers.
At inference time, they optimize a fixed inner loop loss on the observed context, storing long-range information in temporary parameter updates rather than in a full KV cache.
They perform only one gradient step for distilling each context chunk.
By contrast, our method uses a learned recurrent forward pass as the memory-update rule, allowing more flexible forms of consolidation that need not correspond to a one-step gradient descent on a fixed scalar objective.
\citet{tandon2025end} primarily evaluates perplexity on general web-text data, where retrieval and reasoning demands are entangled; we instead use synthetic tasks that independently control reasoning depth and problem length, showing that additional sleep-time computation is most beneficial when reasoning depth increases.
\citet{zhang2026training} attach a LoRA adapter that updates model weights from the current context chunk and evaluate this approach in a reinforcement learning (RL) setting.
Unlike ours, their method updates the weights only once per chunk.

\textbf{Depth-recurrent models.}
Increasing the depth of language models is known to increase their expressivity \citep{merrill2025little}.
Depth-recurrence, is one way to increase depth in transformer models and is one method to make them Turing complete \citep{dehghani2018universal}.
Moreover, the depth of these models can be adaptive \citep{graves2016adaptive,elbayad2019depth,schwarzschild2021can,bansal2022end}.
Recent work has scaled these depth-adaptive language models to large scales, both training from scratch \citep{geipingscaling,zhu2025scaling} and as a post-training objective \citep{mcleish2025teaching}.
Detailed analyses of how best to train depth recurrent models suggest the recurrent depth should be scaled with training compute \citep{prairie2026parcae,schwethelm2026much}.

\textbf{Offline planning.}
Successful planning in structured environments often requires combining newly-observed information with memories of earlier states.
A longstanding view is that animals perform this integration online at choice time~\citep{tolman1948cognitive,momennejad2018offline}.
However, integrating distant memories at choice time can be time-consuming, and offline planning during off-task rest can amortize such cost~\citep{momennejad2018offline}.
Consistent with this view, \citet{momennejad2018offline} show that neural evidence of offline replay during rest predicts improved planning performance for human subjects.
Recent work from the machine learning community studies related mechanisms with artificial neural networks.
\citet{lin2025sleep} propose scaling offline compute by letting LLMs generate expected questions from users and precompute quantities needed to solve them.
\citet{chalvidal2022meta} train a single-layer network on RL environments and show that recursive Hebbian-like weight updates support fast adaptation.
In this paper, we show that recursively updating fast weights during a sleep-like offline phase improves reasoning over evicted context while preserving a strict prediction-phase latency constraint.

\textbf{Sleep.}
Several machine learning methods draw inspiration from biological sleep. These include RL methods in which agents train on model-generated trajectories~\citep{hafner2019dream,sutton1991dyna,ha2018world} or replay
buffers~\citep{mnih2015human}.
Wake-sleep and contrastive divergence methods also use a sleep analogy for their offline phases to train generative models~\citep{hinton1995wake,carreira2005contrastive}.
\citet{lin2025sleep} frames its offline planning phase as ``sleep.''
\citet{behrouzlanguage} study a sleep-inspired memory consolidation method for language models based on RL, parameter expansion, teacher-student distillation, and synthetic data generation.\footnote{This OpenReview
submission has a title that overlaps with our  original title, namely ``Language Models Need Sleep.'' We were unaware of it when choosing our title and have updated ours to avoid confusion.}

\section{Preliminaries}
\subsection{Sequence mixers}
\textbf{Attention.}
Softmax attention~\citep{vaswani2017attention} is a sequence-mixing operation in which each token retrieves information from previous tokens according to query-key similarity.
For the token representation $\x_t$ at timestep $t$, define
\begin{align}
\q_t &= \WQ \x_t, &
\kvec_t &= \WK \x_t, &
\v_t &= \WV \x_t,
\end{align}
where $\q_t,\kvec_t,\v_t \in\R^d$ 
are column vectors, and $\WQ,\WK,\WV$ are learned projection matrices with compatible shapes.
Self-attention stores all previous keys $\kvec_t$ and values $\v_t$ in $\K_t=[\kvec_1,\ldots,\kvec_t]^\top\in\R^{t\times d}$ and $\V_t=[\v_1,\ldots,\v_t]^\top\in\R^{t\times d}$, then computes
\begin{align}
\ovec_t
&=
\V_t^\top
\softmax\!\left(\frac{\K_t \q_t}{\sqrt d}\right).
\end{align}
This allows $\x_t$ to attend to any previous token, but requires storing $\K_t$ and $\V_t$, the KV cache,  whose size grows linearly with sequence length.

\textbf{Linear recurrent layers.}
By contrast, linear recurrent layers, including many SSM-style architectures, store the past in a fixed-size fast-weight state.
A simple Mamba2-style~\cite{dao2024transformers} update can be written as a gated Hebbian-like outer-product rule~\citep{hebb2005organization,schlag2021linear}:
\begin{align}
\Sb_t
&=
\alpha_t \Sb_{t-1} + \beta_t \v_t \kvec_t^\top,
&
\ovec_t
&=
\Sb_t \q_t .
\label{eq:ssm}
\end{align}
Here $\alpha_t\in(0,1)$ is a data-dependent forget gate and $\beta_t\in(0,1)$ is a data-dependent input gate, both computed from $\x_t$.
Unlike the KV cache $\K_t$ and $\V_t$, the fast-weight $\Sb_t$ does not grow in size with $t$.
This makes linear recurrent layers more memory-efficient, but also more lossy: past tokens must be compressed into a fixed-size weight-based memory.
In our experiments we use Gated Delta Networks (GDNs), which add a delta-rule correction to this update; 
however, the specific update rule does not matter for our discussion.

In a language model, a sequence-mixing layer is combined with normalization, residual connections, and an MLP layer to form a block.
We write $\mathcal{B}^{\mathrm{attn}}_\ell$ for a block whose sequence-mixing layer is attention, and $\mathcal{B}^{\mathrm{ssm}}_\ell$ for a block whose sequence-mixing layer is a linear recurrent layer.

For example, an attention-only language model is formed by stacking attention blocks $\modelDepth$ times between an embedding layer and an output projection:
\begin{align}
\mathrm{Embed}
\rightarrow
\mathcal{B}^{\mathrm{attn}}_0
\rightarrow
\cdots
\rightarrow
\mathcal{B}^{\mathrm{attn}}_\ell
\rightarrow
\mathcal{B}^{\mathrm{attn}}_{\ell+1}
\rightarrow
\cdots
\rightarrow
\mathcal{B}^{\mathrm{attn}}_{\modelDepth-1}
\rightarrow
\mathrm{OutProj}.
\end{align}

\textbf{Hybrid models.}
Recent hybrid sequence models~\cite{ren2024samba,dong2024hymba,de2024griffin,arora2024simple} mitigate the cost of self-attention layers by interleaving them with SSM blocks~\cite{yang2024gated,dao2024transformers,schlag2021linear} with fixed-size fast-weight memories. For example:
\begin{align}
  \mathrm{Embed}
  \rightarrow
  \mathcal{B}^{\mathrm{attn}}_0
  \rightarrow
  \mathcal{B}^{\mathrm{ssm}}_1
  \rightarrow
  \mathcal{B}^{\mathrm{attn}}_2
  \rightarrow
  \mathcal{B}^{\mathrm{ssm}}_3
  \rightarrow
  \cdots
  \rightarrow
  \mathcal{B}^{\mathrm{attn}}_{\modelDepth-1}
  \rightarrow
  \mathrm{OutProj}.
\end{align}

\subsection{Synthetic reasoning tasks}
To begin, we study two synthetic tasks to understand our changes in a controlled setting.

\textbf{Rule 110.}
Rule 110~\cite{cook2004universality} is a simple one-dimensional binary cellular automaton that evolves a binary string according to a fixed local transition rule.
The general problem of predicting Rule 110 after $\rolloutStep$ steps is P-complete~\citep{neary2006p}, and no efficient general parallel shortcut is known.
Training a neural network to predict the $\rolloutStep$-th state is therefore a good test to see if the model can carry out deep sequential computation.

\textbf{Depo.}
Depo is a multi-hop knowledge retrieval task introduced by \citet{allen2026physics} to evaluate reasoning depth of a language model.
Each sequence consists of a shuffled directed cycle followed by queries; each query asks for the node reached after $\hopCount$ outgoing edges from a start node, with larger $\hopCount$ requiring deeper graph traversal.

These tasks allow us to vary reasoning demand while holding sequence length fixed, isolating a model's reasoning capability from its information retrieval capability.

\section{Motivating example: Can attention-SSM hybrid models reason about context they can no longer attend to?}
\label{sec:motivating}

Attention-SSM hybrid models are often motivated by the idea that fast-weight memory can compensate for limited attention windows \cite{ren2024samba}, compressing information from past tokens once they are no longer directly accessible.
In this section, we explore a case where this hybrid mechanism fails.

Consider the following example drawing on cellular automaton Rule 110~\cite{cook2004universality}.
In this setting, we train the model on four independent length-24 binary strings, each representing an initial state for Rule 110.
Here, we use a character-level tokenizer (i.e., `0' and `1' define tokens).
The four states are unrelated to each other (i.e., they are not obtained by unrolling the previous state).
After processing the all four binary strings of length $\sequenceLen:= 24\times 4=96$, the model must later predict the first bit of each state after $\rolloutStep$ transitions.
Since there are four label tokens following the states, the total sequence length $\sequenceLen$ is $100$.
An example sequence is:
\begin{center}
  \begin{tabular}{@{}c@{\;}c@{\;}c@{\;}c@{\;}c@{\;}c@{\;}c@{\;}c@{\quad}c@{\quad}c@{\quad}c@{\quad}c@{}}
    $\overbrace{\texttt{0101}\ldots\texttt{1101}}^{24\ \text{bits}}$ &
    \textcolor{red}{$\vert$} &
    \texttt{1101}\ldots\texttt{1000} &
    \textcolor{red}{$\vert$} &
    \texttt{1101}\ldots\texttt{0110} &
    \textcolor{red}{$\vert$} &
    \texttt{0011}\ldots\texttt{0110} &
    \textcolor{red}{$\vert$} &
    \textcolor{red}{\texttt{1}} &
    \textcolor{red}{\texttt{0}} &
    \textcolor{red}{\texttt{1}} &
    \textcolor{red}{\texttt{0}} \\
    state0 & & state1 & & state2 & & state3 & & label0 & label1 & label2 & label3
  \end{tabular}
\end{center}
The first answer token \textcolor{red}{1} (label0) is obtained by unrolling 0101...1101 (state0) $\rolloutStep$ times and taking the first bit from it, and so on.
$\rolloutStep$ controls the reasoning depth required to solve this task: when $t=0$ (no rollout), this becomes a simple first-bit retrieval task, and the task becomes more difficult as $t$ increases.

To stress-test whether SSM can complement self-attention by providing past information, 
we impose a strict context window size as well as a \textit{hard-eviction constraint}:
we clear the context window every \(24\) tokens, and we denote this with $\windowSize=24$.
This means that the model can only see one state in context at a time and must fully encode this information into its fast weights $\Sb_t$, as the KV cache $\K_t$ and $\V_t$ are \textit{fully evicted} before moving onto the next state.
The hard eviction boundary is denoted by \textcolor{red}{$\vert$}.

This hard eviction constraint naturally divides a sequence into two distinct phases:
\begin{itemize}
    \item the \textbf{consolidation phase} (the first 96 tokens in the example sequence), during which the model must encode context into its fast weights $\Sb_t$; and
    \item the \textbf{prediction phase} (the last 4 tokens in the example sequence), during which the model predicts the answer tokens.
\end{itemize}
We impose a \textit{prediction-phase latency constraint}: during the prediction phase, each answer token is predicted with a single standard forward pass. Extra loops or chain-of-thought tokens are disallowed because they increase prediction latency.
Thus, all information needed to predict the labels must already have been consolidated into the fast weights before the prediction phase begins.

Under this hard eviction constraint, a standard transformer cannot do better than random guessing as the KV cache has been destroyed before prediction is made.
SSMs or attention-SSM hybrid models can do better 
than random guessing because they can store the initial states in their fast weights.
For example, one way to solve this task is to simulate the $\rolloutStep$-step state evolution once the context is full, store the first bit of each evolved state in the fast weights, and retrieve this bit at prediction time.
However, \Cref{fig:automata-baseline} shows that the performance of a 4-layer GDN-attention hybrid model (with an attention $\rightarrow$ GDN $\rightarrow$ attention $\rightarrow$ GDN layout) drops rapidly as $\rolloutStep$ increases.
This drop is not due to the memory-capacity limitation identified in prior work~\cite{jelassi2024repeat,arora2024simple}: we vary only $\rolloutStep$ while keeping the sequence length $\sequenceLen$ fixed.
Instead, the difficulty comes from the deep sequential computation needed to simulate the automaton for $\rolloutStep$ steps, which a fixed-depth model cannot scale with.

\textbf{On task failures.}
When we say that a model fails or degrades on a task, we do not mean that the architecture could never learn the task with unlimited data, compute, or training time.
Our claims concern performance under a fixed training-token budget.
This budgeted setting matters because reasoning-intensive data is sparse even in web-scale corpora.
Budget-controlled synthetic tasks can expose trends that align with phenomena observed in larger-scale pretraining earlier and more clearly~\citep{allen2026physics}.

\section{LLM Sleep: Offline Recursive Memory Consolidation}
\label{sec:method}

Now, we introduce a solution to the above example: 
we introduce a \emph{sleep} during LLM training, in which the model performs recursion during a consolidation phase, before evicting tokens from attention layers once the context window is full.
In this way, we can scale compute to handle deep reasoning tasks (e.g., a large $\rolloutStep$ from our motivating example) while still obeying a prediction-phase latency constraint.
For example, if we loop over all  $\modelDepth$ blocks, it looks like:

\begin{align}
\mathrm{Embed}
\rightarrow
\left[
\mathcal{B}^{\mathrm{attn}}_0
\rightarrow
\mathcal{B}^{\mathrm{ssm}}_1
\rightarrow
\cdots
\rightarrow
\mathcal{B}^{\mathrm{attn}}_{\modelDepth-1}
\right]^{\times \sleepPasses}
\rightarrow
\mathrm{OutProj}
\label{eq:ssm-sleep}
\end{align}
where the superscript $\times N$ denotes $N$ looped passes over the architecture. 
\begin{algorithm}[!htbp]
  \caption{Our LLM sleep training with hard eviction.}
  \label{alg:sleep}
  \small
  \begin{algorithmic}[1]
  \Require tokens $x$, loss mask $m$, window size $\windowSize$, sleep passes $\sleepPasses$
  \State \textbf{Zero-initialize} SSM fast weights $\Sb$
  \State Split $x,m$ into non-overlapping chunks of length at most $\windowSize$
  \For{each token chunk $c$ and its loss mask $m_c$}
      \State $h \gets \mathrm{Embed}(c)$
      \If{$m_c$ is all-zero} \Comment{consolidation phase}
          \For{$n = 1,\ldots,\sleepPasses$}
              \State $h,\Sb \gets \mathrm{Blocks}(h,\Sb)$ 
          \EndFor
      \Else \Comment{prediction phase}
          \State $h,\Sb \gets \mathrm{Blocks}(h,\Sb)$ 
          \State $\mathcal{L} \gets \mathrm{MaskedCE}(\mathrm{OutProj}(h), c, m_c)$ \Comment{Masked cross entropy loss}
      \EndIf
  \EndFor
  \State Backpropagate $\mathcal{L}$ and take an optimizer step
  \end{algorithmic}
\end{algorithm}

\begin{figure}[ht]
  \centering
  \includegraphics[width=0.8\textwidth]{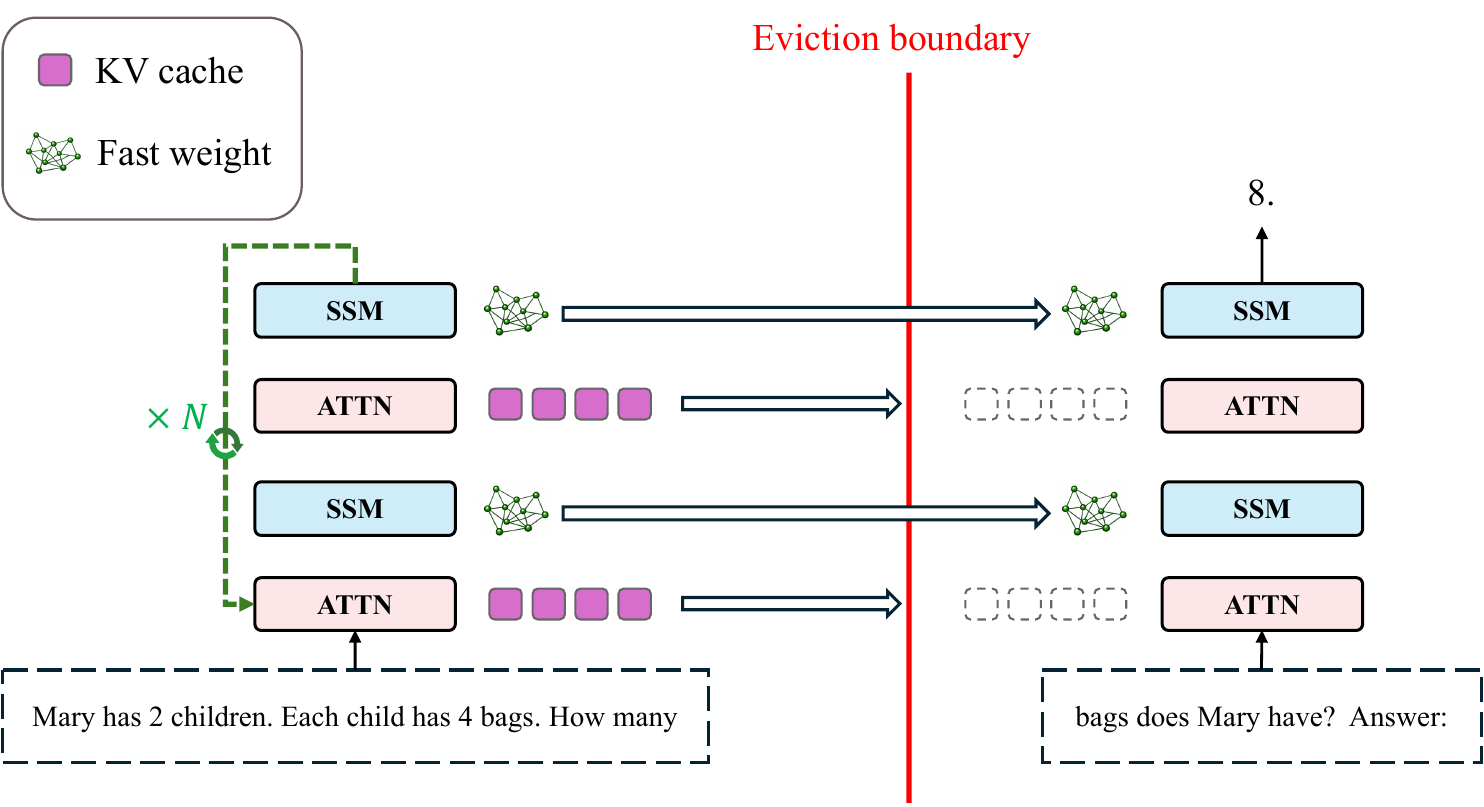}
  \caption{At the eviction boundary, an SSM-attention hybrid performs $\sleepPasses$ offline recurrent passes over the current context before discarding the attention cache.
  These recurrent passes update the fast weights in the SSM blocks, allowing later predictions to use consolidated context without wake-time looping. }
  \label{fig:method}
\end{figure}

\Cref{fig:method} describes the architecture in detail.
We initialize from an SSM-attention hybrid model with a fixed context-window size $\windowSize$, where the attention cache is fully evicted every $\windowSize$ tokens.
Before evicting the KV cache every $\windowSize$ tokens, the model performs $\sleepPasses$ recurrent passes to iteratively update the fast weights inside the SSM blocks following \Cref{eq:ssm}; with $\sleepPasses=1$, it reduces to a vanilla SSM-attention hybrid model.
We call the phase when the model is iteratively updating the fast weights a \textbf{sleep}.

After recurrently refining the fast weights, the KV cache is evicted and the next $\windowSize$ tokens are processed.
After processing the full context, the model predicts the answer based on the refined memory and current context \textit{in a single forward pass}.
The model is trained to minimize the prediction error by backpropagating through the entire computational graph shown in \Cref{eq:ssm-sleep}, similarly to other depth-recurrent models~\cite{dehghani2018universal,graves2016adaptive}.
Unlike prior depth-recurrent models where gradient flows through recursively refined feature vectors, the gradient flows through the refined fast weights because we discard the refined features after sleep.
\Cref{alg:sleep} summarizes the training procedure.

\FloatBarrier
\section{Experiments}
\label{sec:experiments}
Our experiments test whether longer sleep, implemented by increasing $\sleepPasses$, produces fast weights that support deeper reasoning over states that are no longer present in the attention cache.
This requires more than storing evicted tokens: the model must encode past context into fast weights $(\Sb_t)$ in a form that supports nontrivial computation after the cache has been cleared, while still using only a single forward pass at prediction time.
We evaluate this question across increasingly more difficult settings.
First, the cellular automaton task varies the rollout step $\rolloutStep$, isolating the depth of reasoning required over each evicted state.
First, the Depo task \cite{allen2026physics} adds a harder compression problem: the model must encode a fragmented graph into fast weights and later answer unseen multi-hop queries over it.
Finally, we consider GSM-Infinite~\cite{zhou2025gsm}, where we fine-tune the pre-trained Jet-Nemotron 2B~\cite{gu2025jet} and Ouro 1.4B~\cite{zhu2025scaling} on a synthetic math-reasoning dataset.

\textbf{Experiment details.}
Following \citet{mcleish2025teaching}, we use the Muon optimizer for all experiments.
We fix the AdamW learning rate to $5\mathrm{e}{-}5$ and tune only the Muon learning rate.
For \Cref{sec:motivating} and \Cref{sec:exp-cellular-automaton}, we use a 4-layer GDN-attention hybrid model with hidden dimension $d=256$.
We tune the Muon learning rate on the $\sleepPasses=1$ model, giving the no-loop baseline an advantage, and use the selected value, $2\mathrm{e}{-}
3$, for all looped models.
For \Cref{sec:exp-depo}, we use the Jet-Nemotron architecture~\citep{gu2025jet}, an SSM-attention hybrid model fine-tuned from Qwen 2.5 1.5B by
replacing some attention layers with Jet layers, which use dynamic convolution instead of the fixed convolution in GDN.
To roughly match the small model size in \citet{allen2026physics}, we train a 10-layer model from scratch with hidden dimension $d=512$.
We apply the same tuning protocol as above: tune the Muon learning rate on the $\sleepPasses=1$ baseline and use $2\mathrm{e}{-}3$ for the looped
models.
For \Cref{sec:exp-gsm-infinite}, we use pre-trained Ouro 1.4B~\citep{zhu2025scaling} and Jet-Nemotron 2B~\citep{gu2025jet} models, and set the Muon
learning rate to $1\mathrm{e}{-}3$ following \citet{mcleish2025teaching}.
The automaton experiments require less than one A6000 GPU-day.
The Depo and GSM-Infinite experiments require roughly 1--2 H100 GPU-days per run.
For the batch size, we use 512 for automaton, 128 for Depo, and 256 for GSM-Infinite.
For fair comparison, we fix random seeds ensuring that all runs use exactly the same data ordering.

\subsection{Task: Cellular automaton}
\label{sec:exp-cellular-automaton}
\begin{figure}[t!]
  \centering
  \begin{subfigure}[b]{0.48\textwidth}
    \centering
    \includegraphics[height=1.45in]{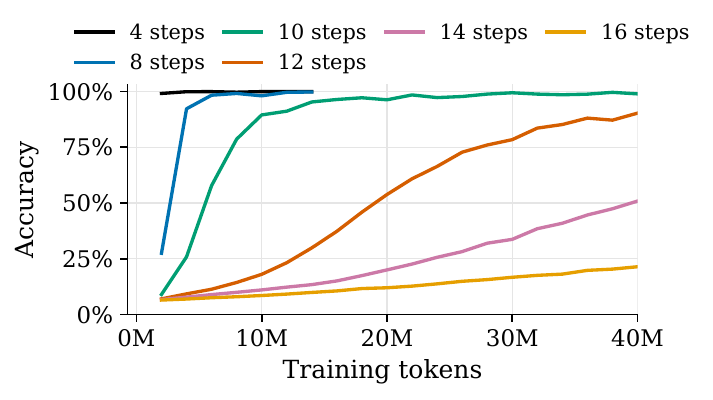}
    \caption{Effect of rollout step $\rolloutStep$.}
    \label{fig:automata-baseline}
  \end{subfigure}
  \begin{subfigure}[b]{0.48\textwidth}
    \centering
    \includegraphics[height=1.45in]{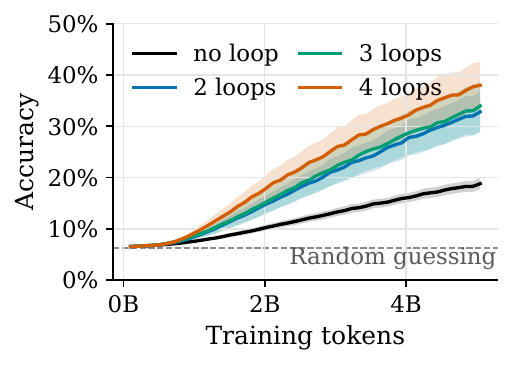}
    \caption{Effect of offline looping for $\rolloutStep=32$.}
    \label{fig:automata-looped}
  \end{subfigure}
  \caption{\textbf{Increasing $\sleepPasses$ improves performance on cellular automaton.} \textbf{Left:} 
  Each curve represents a different number of rollout steps $\rolloutStep$ for a hybrid attention-SSM architecture, 
  as in the motivating example section. Increasing $t$ makes the task harder for a vanilla attention-GDN hybrid model. We early-stop 4- and 8-step runs as they converge earlier.
  \textbf{Right:} For a challenging reasoning task ($\rolloutStep=32$), additional offline sleep loops improve accuracy while preserving single-pass wake-time prediction. Here, lines show the mean over seven random seeds, and shaded area shows standard error.
  }
  \label{fig:automata}
\end{figure}

In \Cref{sec:motivating}, we see how vanilla SSM-attention hybrid models fail 
on the automaton task when $t$ is large, as hybrid models cannot scale compute when performing memory consolidation.
In \Cref{fig:automata-looped}, we use the same architecture from \Cref{fig:automata-baseline}: a 4-layer GDN-attention hybrid model, with an attention $\rightarrow$ GDN $\rightarrow$ attention $\rightarrow$ GDN layout. 
Our method additionally uses the `sleep' during the consolidation phase discussed in \Cref{sec:method}, where we use recurrence to iteratively update the fast weights. We study using 2 to 4 recurrent updates here.

We train this looped hybrid architecture on a setting that requires substantial reasoning compute and is challenging for the non-recurrent architecture: $\rolloutStep=32$.
In \Cref{fig:automata-looped}, ``2 loops'', ``3 loops'', and ``4 loop'' mean the model uses a \textit{sleep} for memory consolidation, while "no loop" is the baseline.
\Cref{fig:automata-looped} shows that the non-looped model remains close to random guessing, reaching only about $10\%$ exact accuracy after nearly $5$B training tokens.
Adding offline passes improves both learning speed and final accuracy under the same token budget: two loops achieves approximately $20\%$ accuracy, while three and four loops achieve above $30\%$.
Because the context length, eviction rule, and prediction-phase computation are fixed across these runs, the improvement comes from additional consolidation-time computation during sleep.

\subsection{Task: Depo}
\label{sec:exp-depo}
Next, we evaluate Depo, the $\hopCount$-hop knowledge retrieval task introduced by \citet{allen2026physics}.
Each sequence consists of a shuffled directed cycle followed by queries; each query asks for the node reached after $\hopCount$ outgoing edges from a start node, with larger $\hopCount$ requiring deeper graph traversal.
\citet{allen2026physics} show that SSMs perform substantially worse than transformers on this task, despite having enough fast weight capacity to store the context.
This suggests that the bottleneck is not storage alone, but organizing stored edges into a representation that supports later multi-hop retrieval~\cite{noroozizadeh2025deep}.
An example sequence from Depo is:

\begin{center}
$\overbrace{
\mbox{\texttt{b->a, f->l, ...}
\textcolor{red}{$\vert$}
\texttt{...}
\textcolor{red}{$\vert$}
\texttt{..., e->b}}
}^{\text{shuffled directed cycle}}$
\textcolor{red}{$\vert$}
$\overbrace{
\mbox{\texttt{1 hop after a:} \textcolor{red}{\texttt{c}}
\quad \texttt{...} \quad
\texttt{4 hops after e:} \textcolor{red}{\texttt{d}}}
}^{\text{query and answer}}$
\end{center}

Here \textcolor{red}{$\vert$} denotes an eviction boundary, and red text denotes answer tokens.

In our setting, each cycle contains up to 75 nodes and spans up to 300 tokens; shorter instances are left-padded to 300 tokens both at test and train time.
The query-answer portion then follows, with 10 query-answer pairs spanning up to 60 tokens, making the total sequence length $\sequenceLen=360$.
The model's window size is $\windowSize=75$, so each cycle is fragmented across four cache windows.
When the model predicts the query answers, the cycle context has been evicted from the KV cache.
Depo is harder than the cellular automaton task for two reasons.
First, each cycle is fragmented across four cache windows, whereas each automaton state fits within a single window.
Second, the model must form a query-agnostic representation because both $\hopCount$ and the start node are randomly sampled for each example, whereas $\rolloutStep$ is fixed in the automaton task.

In Depo, $\hopCount$ controls task difficulty: larger $\hopCount$ makes the query more difficult because the model must perform longer multi-hop
  traversal to recover the answer.
Following \citet{allen2026physics}, we uniformly sample $\hopCount$ from $[1, 16]$ during training and measure test loss on held-out examples with $\hopCount=\{1, 2, 4, 8, 16\}$.

\begin{figure}[t!]
  \centering
  \includegraphics[width=0.8\textwidth]{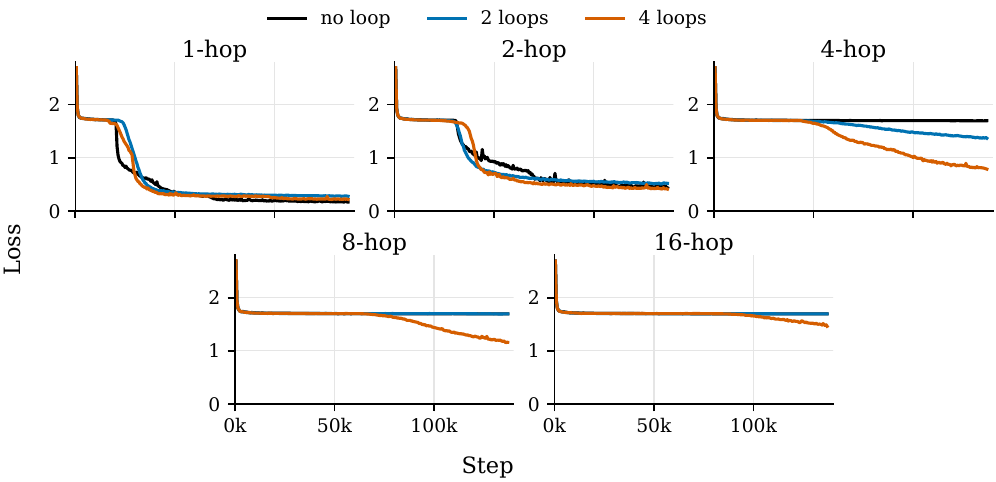}
  \caption{\textbf{Increasing $\sleepPasses$ improves performance on Depo.} Test loss of a 4-layer GDN-attention hybrid on the $\hopCount$-hop knowledge retrieval task. Additional offline loops accelerate learning, especially for more reasoning-intensive, higher-hop queries.}
  \label{fig:depo}
\end{figure}

\Cref{fig:depo} shows test loss on held-out examples over training steps, with each subplot corresponding to a hop count $\hopCount\in\{1,2,4,8,16\}$ and each curve comparing a model with $\sleepPasses\in\{1,2,4\}$ offline loops.
We see that increasing the number of offline loops improves learning speed
for queries that require 4 or more hops.
The 1-loop model makes little progress on 4-hop and harder queries, and the 2-loop model similarly stalls on 8-hop and harder queries.
Within our training budget, only the 4-loop model begins to improve on the hardest 16-hop task.

\subsection{Task: GSM-Infinite}
\label{sec:exp-gsm-infinite}
To test whether the trend from the controlled tasks extends to pretrained LLMs, we evaluate on GSM-Infinite~\cite{zhou2025gsm}, a synthetic reasoning benchmark modeled after GSM8K~\cite{cobbe2021training}.
GSM-Infinite is still structured enough for controlled analysis, but realistic enough that training on it can improve a model's reasoning capabilities on other tasks~\cite{kabra2026learning}.
As GSM-Infintie is procedurally generated we can generate distinct training and evaluation datasets from the same distribution, similarly to \citet{kabra2026learning}.
Our evaluation set is 1,600 held-out examples.
The dataset controls problem length by adding distractor tokens that resemble the rest of the problem, making them difficult to ignore, and controls difficulty by varying the number of arithmetic operations required to solve the problem.
Unlike retrieval-focused long-context tasks such as RULER~\cite{hsieh2024ruler}, simple retrieval-augmented baselines fail~\cite{zhou2025gsm} on GSM-Infinite, indicating that the task requires both long-context processing \textit{and} multi-step reasoning.
GSM-Infinite is challenging even for reasoning-optimized frontier models, whose accuracy decays as the number of required operations increases~\citep{zhou2025gsm}.

In our experiments, each problem contains between 2,000 and 3,300 tokens, and the number of operations is sampled uniformly from $[1,8]$.
We place the question before the context and exclude Chain-of-Thought traces from the data, forcing the model to the final answer in the single prediction time forward pass alone.
This order gives the model the query before it reads the long problem context, allowing it to selectively consolidate information relevant to the question while ignoring filler tokens.
We set the model's context-window size to $\windowSize=2000$, so a full problem does not fit in the active context window and the model cannot attend to a majority of the problem context at prediction time.

There are two complementary ways of instantiating our method from a pre-trained model: starting from an SSM-attention hybrid and fine-tuning it with sleep time recurrence, or starting from a depth-recurrent model and adding SSM memory layers.
We explore both, fine-tuning the hybrid Jet-Nemotron 2B~\cite{gu2025jet}, and the recurrent Ouro 1.4B~\citep{zhu2025scaling}.
Jet-Nemotron is an SSM-attention hybrid model fine-tuned from Qwen 2.5 1.5B by replacing some attention layers with Jet layers, which use dynamic convolution instead of the fixed convolution in GDN.
Ouro is a looped attention-only model, so we insert 6 Jet layers without MLP layers to augment Ouro with fast weight memory while increasing the total parameter count by less than 10\%.

For Jet, we loop over the middle 14 blocks out of the total 28 blocks.
Looping over middle-blocks only is a common practice in depth-recurrence models \cite{mcleish2025teaching,geipingscaling}.
For Ouro, we loop over the entire blocks following how the model is pre-trained \cite{zhu2025scaling}.
To keep memory cost during training manageable while using a reasonable batch size, we use $N=\{1,2,4\}$ for Ouro. 
Since Jet loops over only a half of the entire blocks, we use $\{1,2,4,6\}$.

  \begin{figure}[t!]
    \centering
    \begin{subfigure}[t]{0.49\textwidth}
      \centering
      \includegraphics[height=2.1in, keepaspectratio]{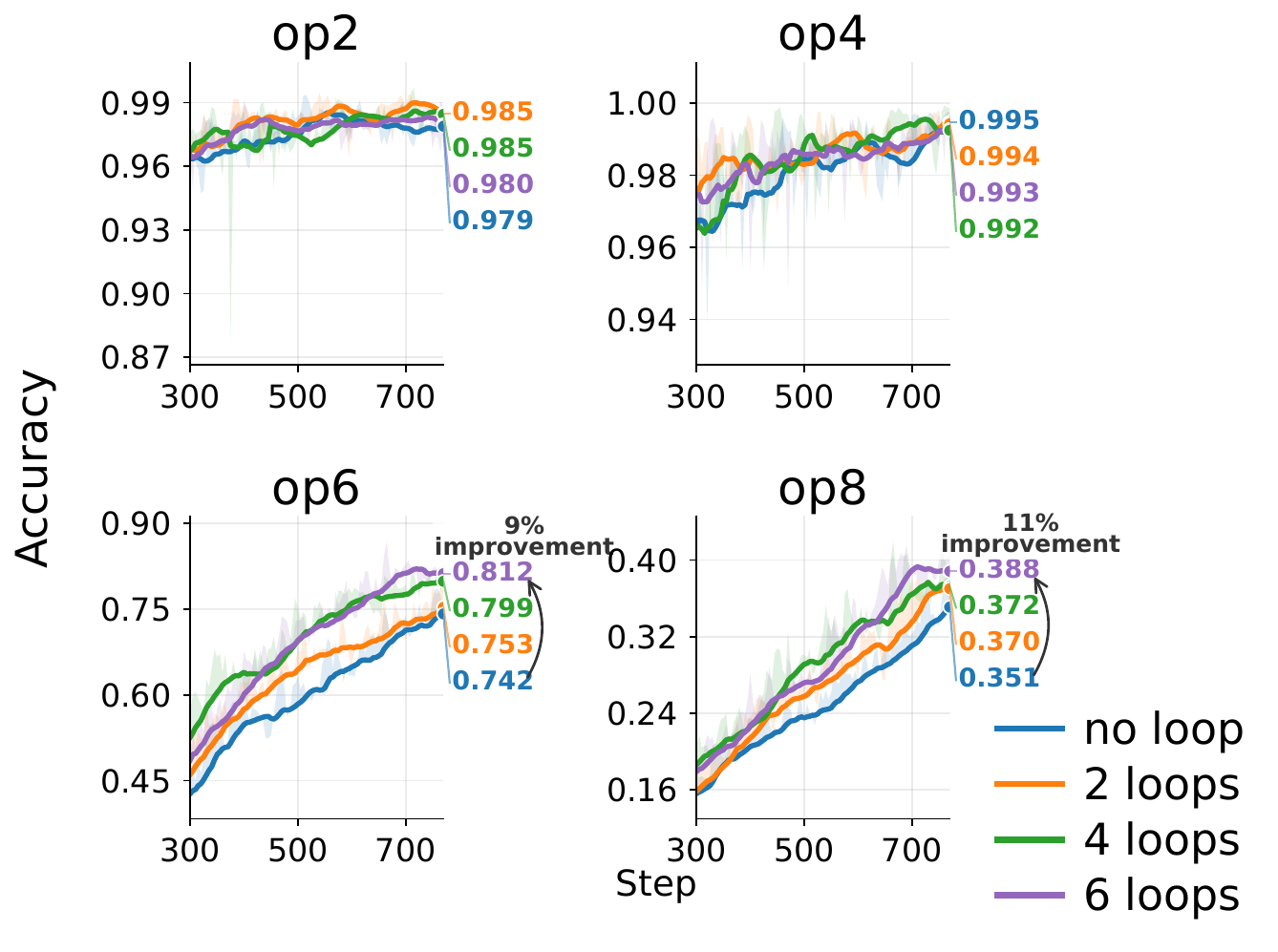}
      \caption{Jet-Nemotron 2B.}
      \label{fig:gsm-infinite-jet}
    \end{subfigure}
    \hfill
    \begin{subfigure}[t]{0.49\textwidth}
      \centering
      \includegraphics[height=2.1in, keepaspectratio]{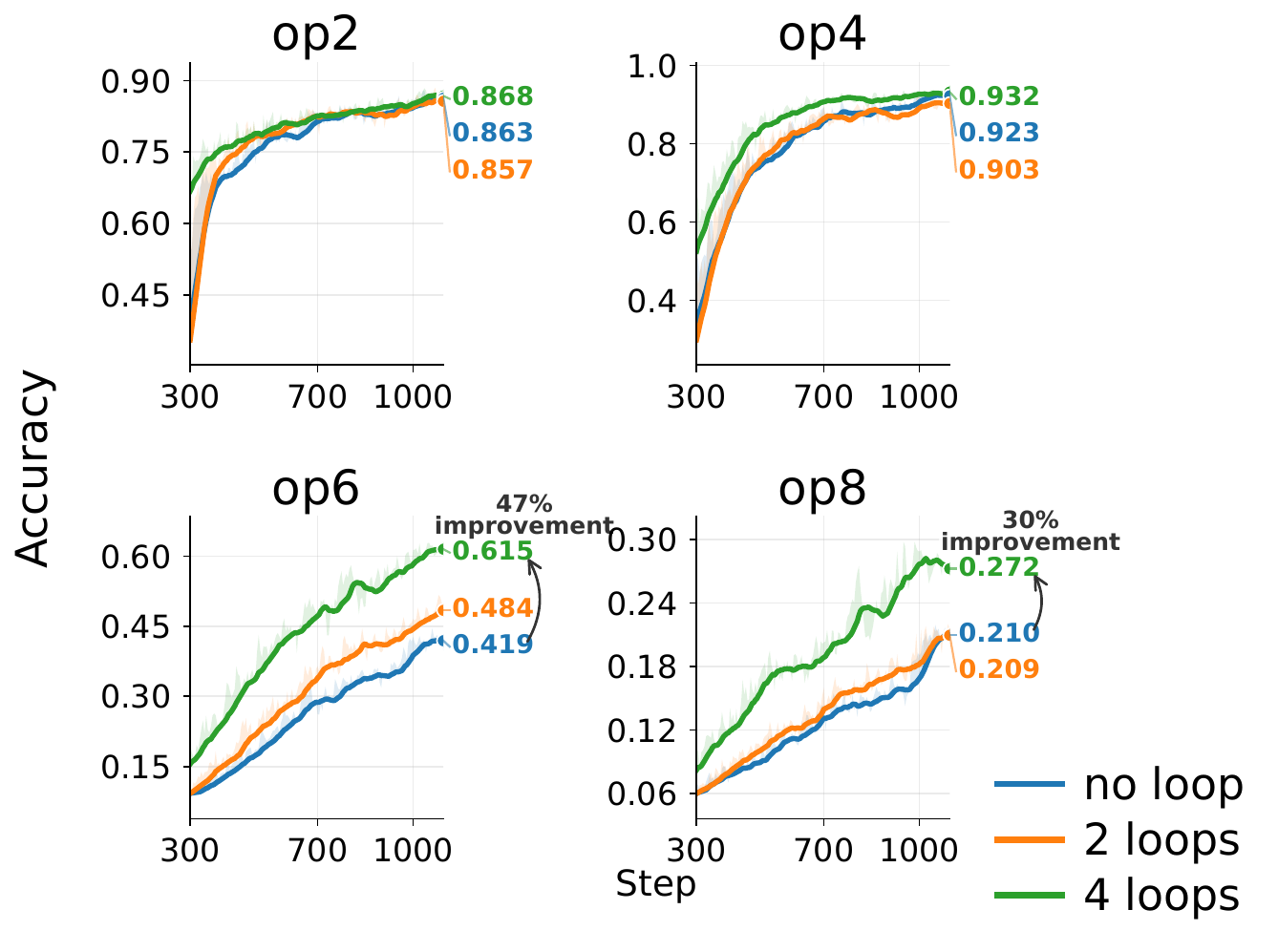}
      \caption{Ouro 1.4B.}
      \label{fig:gsm-infinite-ouro}
    \end{subfigure}
    \caption{\textbf{Increasing $\sleepPasses$ improves performance on GSM-Infinite.} GSM-Infinite accuracy over training steps. Subplots group examples by the number of arithmetic operations required by the problem, and colors indicate the number of offline loops $\sleepPasses$ used before cache eviction.
  Additional loops improve accuracy most clearly on harder problems with more operations, where single-loop models have less sleep-time computation available to organize the evicted context into useful fast weights.
  }
    \label{fig:gsm-infinite}
  \end{figure}

\Cref{fig:gsm-infinite} shows the accuracy trend over training steps, with each subplot corresponding to a different number of operations required to solve the problem, ranging from 2 to 8.
We see that the trend from the pretraining from scratch experiments persists in a more realistic math-reasoning setting.
For easier two- and four-operation problems, accuracy often approaches saturation regardless of the number of loops, especially for Jet, which has more fast weight memory capacity than Ouro.
However, as the number of required operations increases, the gap between loop counts widens: additional offline recurrence improves both final accuracy and learning speed on the six- and eight-operation settings.
For Jet, six loops improves final accuracy on six-operation problems from $0.742$ to $0.812$ and on eight-operation problems from $0.351$ to $0.388$.
For Ouro, four loops improves final accuracy from $0.419$ to $0.615$ on six-operation problems and from $0.210$ to $0.272$ on eight-operation problems.
The gap is wider for Ouro, which may reflect its depth-recurrent pretraining.
These results suggest that sleep-time computation can support multi-step reasoning even on realistic math-reasoning data and with pre-trained LLMs.
We report mean accuracy and standard error across three random seeds for Ouro after 815 training steps in \Cref{tab:gsm-infinite-seed-variability}; we found that run-to-run variance is small.

\subsection{Sliding-window eviction}
\label{sec:exp-sliding-window-eviction}
So far, we have assumed that the model's context window is completely evicted whenever it is full.
We can instead use a sliding-window eviction strategy: after sleep, the model retains the most recent $\windowSize-1$ tokens in the attention cache and evicts only older tokens.
This does not increase peak inference-time memory: the active context is still capped at $\windowSize$ tokens, as in sliding-window attention (SWA).
With $\sleepPasses=1$, this reduces to a standard SWA-SSM hybrid model~\citep{ren2024samba}; with $\sleepPasses>1$, the model performs additional recursive consolidation before older context leaves the attention cache.

We evaluate this strategy on GSM-Infinite with $\windowSize=512$, so the total sequence length $\sequenceLen$ is roughly $4$--$6\times$ the window size.
We fine-tune Ouro 1.4B with $\sleepPasses\in\{1,2,4\}$.
Analogously to observations in prior work~\citep{cabannes2025short}, we find that giving the model access to a sliding-window KV cache can make the newly inserted Jet layers underutilized.
We therefore first warm up only the Jet layers for one epoch and then train the full model for two epochs.
This SSM-only warm-up stage is standard when converting attention-only models into attention-SSM hybrids~\citep{wang2024mamba,bick2024transformers,gu2025jet}.
We find that for $\sleepPasses>1$, using hard eviction for the warm-up stage is crucial for the model to learn to refine the fast weights.

\begin{figure}[t!]
  \centering
  \includegraphics[width=1\textwidth]{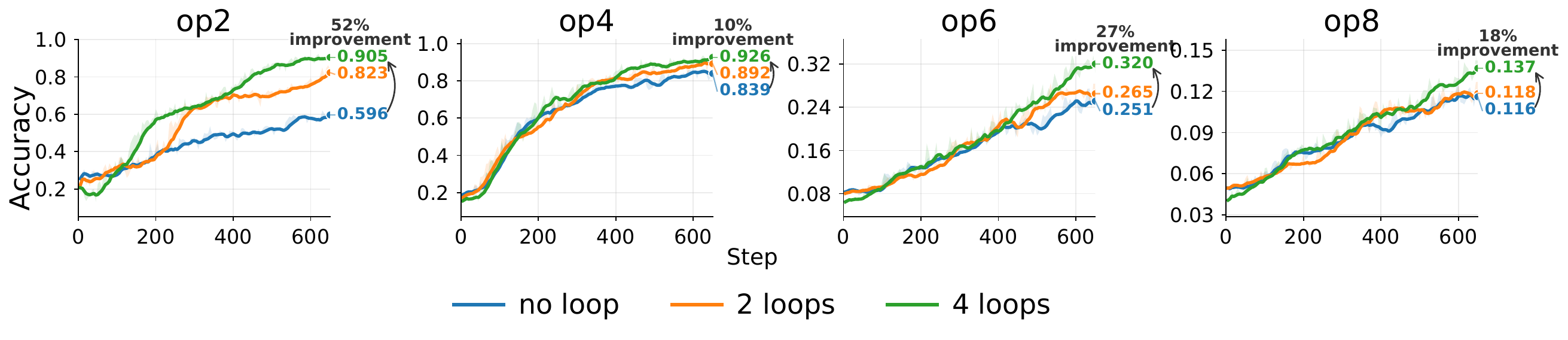}
  \caption{\textbf{Increasing $\sleepPasses$ improves accuracy on GSM-Infinite with sliding-window eviction.} GSM-Infinite accuracy with sliding-window eviction over training steps. We fine-tune Ouro 1.4B with window size $\windowSize=512$ and compare $\sleepPasses\in\{1,2,4\}$ sleep passes.}
  \label{fig:gsm-infinite-sliding-window-eviction}
\end{figure}

\Cref{fig:gsm-infinite-sliding-window-eviction} shows accuracy over training steps, where the curve labeled \texttt{no loop} corresponds to the SWA-SSM hybrid baseline with $\sleepPasses=1$.
Increasing $\sleepPasses$ improves accuracy at all operation counts, matching the trend in \Cref{fig:gsm-infinite}.
Unlike in \Cref{fig:gsm-infinite}, where the window size is $\windowSize=2000$, this baseline performs poorly even on two-operation problems, which are the least reasoning-heavy and therefore more directly stress retrieval under distractor tokens.
On the other hand, using loops drastically improves accuracy from 0.596 to 0.905, an 52\% improvement.
This suggests that when the active attention window is several times smaller than the sequence length, \textbf{longer sleep duration helps not only with multi-step reasoning, but also with compressing and retrieving relevant context.}

\subsection{Training throughput}
\label{sec:throughput}

\begin{figure}[ht]
  \centering
  \begin{subfigure}[b]{0.48\textwidth}
    \centering
    \includegraphics[height=1.45in]{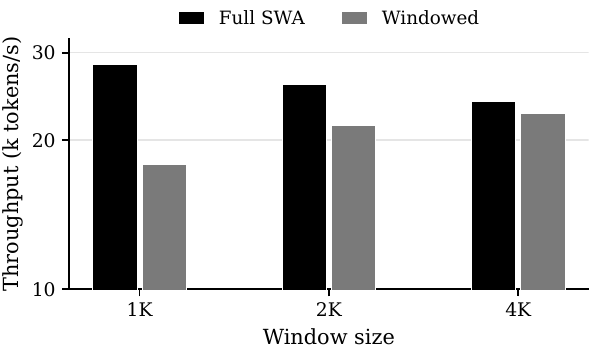}
    \caption{Parallel SWA vs windowed.}
    \label{fig:compute-windowed}
  \end{subfigure}
  \begin{subfigure}[b]{0.48\textwidth}
    \centering
    \includegraphics[height=1.45in]{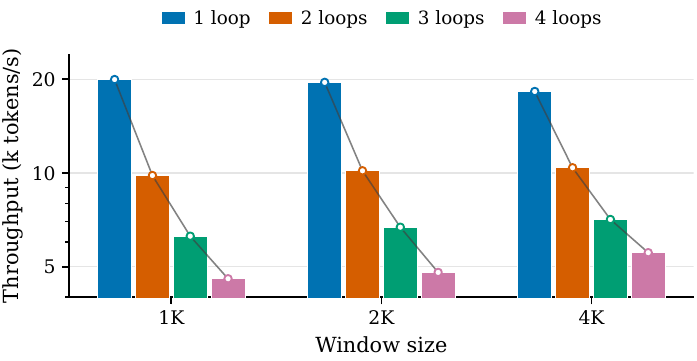}
    \caption{Effect of varying $\sleepPasses$.}
    \label{fig:compute-loop}
  \end{subfigure}
  \caption{\textbf{Recurrence across context windows incur minimal training overhead; recurrent-depth linearly increases cost.} Training throughput comparison on 1 NVIDIA H200 GPU. Sequence length is set to $12,000$. (a) When window size $\windowSize$ is sufficiently large, serialness across context windows do not meaningfully change the throughput compared to the fully parallel baseline. (b) Throughput is roughly inversely proportional to $\sleepPasses$. For each setting, batch size is tuned to optimize the GPU utilization. (b) additionally uses activation checkpointing across context chunk axis to prevent out-of-memory error. FlashAttention 2~\citep{dao2024flashattention} is used.}
  \label{fig:compute}
\end{figure}

Here we analyze how our method affects training throughput in terms of the number of tokens processed per second compared to a SWA-SSM hybrid baseline.
We use Ouro 1.4B model from \Cref{sec:exp-gsm-infinite}.

\textbf{Recurrence across context windows.}
Unlike standard teacher-forced transformer training, which can process all token positions in parallel, our training is recurrent across context windows, since before window $j+1$ can be processed, the model must finish processing window $j$ and perform the $\sleepPasses$ sleep passes that refine the fast weights.
The updated fast weights then become the state used to process window $j+1$, creating a sequential dependency across windows.
This prevents full parallelization along the sequence axis.
However, this loss of sequence-axis parallelism need not hinder wall-clock training time when the window size $\windowSize$ is large enough to keep the GPU saturated, as can occur in long-context training regimes where both $\sequenceLen$ and $\windowSize$ are large, as shown in \Cref{fig:compute-windowed}.

\textbf{Recurrent-depth cost.}
In addition, as in other depth-recurrent models, training cost grows roughly linearly with the number of recurrent steps $\sleepPasses$, as shown in \Cref{fig:compute-loop}.
However, as we see in our experiments, increasing recurrence consistently improves task performance compared to non-recurrent models.

\section{Discussion and Limitations}
\label{sec:limitation}

Our method preserves single-pass prediction-phase latency by moving the extra recurrent computation into the consolidation phase, but this gain is not free: during training, we need to perform $\sleepPasses$ deeper forward and backward passes, which can make training slow and unstable.
Tackling these challenges is an active topic in recurrent-depth training, with possible approaches including implicit gradients~\citep{bai2019deep} and truncated backpropagation through time~\citep{geipingscaling,mcleish2025teaching}, as well as various techniques to stabilize training \citep{prairie2026parcae,geipingscaling}.

Sleep makes training sequential across context and depth dimension, but this sequentiality is also why our method shows gains on the tasks we consider, whose solutions are themselves sequential.
Many reasoning, simulation, and decision-making
problems often targeted by modern machine learning appear to have this property~\citep{liu2025serial}.
Attempting to solve inherently sequential tasks with fully parallel computation encourages brittle shortcut solutions~\citep{liu2022transformers,liu2025serial}.

\section{Conclusion}
\label{sec:conclusion}
We propose a sleep-like process in which a model performs multiple recursive forward passes to iteratively refine its fast weights before evicting the corresponding context from the attention cache.
Unlike vanilla attention-SSM hybrid model, sleep allows models to reason deeply about past context that they can no longer attend to.
Across controlled synthetic tasks and a more realistic mathematical reasoning benchmark, we show that increasing the number of recursions, or sleep duration, improves the model's ability to perform deep sequential computation over evicted context.

\section*{Broader Impact}
\label{sec:broader}
This work studies memory consolidation and reasoning in language models, which are important ingredients for building more capable long-context
systems.
Our contribution is primarily methodological and is evaluated on controlled synthetic tasks and modest-scale pretrained models.
We therefore do not expect the risks to exceed those of other work in this area.

\section*{Acknowledgements}
We gratefully acknowledge Modal for providing generous GPU resources. This work was supported in part by NSF grant CNS-2148359.

\bibliographystyle{plainnat}
\bibliography{reference}

\appendix

\section{Seed variability}

We report mean accuracy and standard error across random seeds for GSM-Infinite in \Cref{tab:gsm-infinite-seed-variability}.

\begin{table}[htbp]
  \centering
  \caption{Mean accuracy and standard error on GSM-Infinite across three random seeds after 815 training steps using Ouro, for different numbers of recursions $\sleepPasses$. Columns group problems by the number of arithmetic operations.}
  \label{tab:gsm-infinite-seed-variability}
  \begin{tabular}{ccccc}
    \toprule
    $\sleepPasses$ & op2 & op4 & op6 & op8 \\
    \midrule
    1 & $0.678 \pm 0.149$ & $0.768 \pm 0.101$ & $0.281 \pm 0.074$ & $0.124 \pm 0.026$ \\
    2 & $0.841 \pm 0.006$ & $0.886 \pm 0.003$ & $0.414 \pm 0.007$ & $0.168 \pm 0.007$ \\
    4 & $0.842 \pm 0.003$ & $0.923 \pm 0.006$ & $0.530 \pm 0.008$ & $0.230 \pm 0.003$ \\
    \bottomrule
  \end{tabular}
\end{table}

\newpage

\end{document}